\documentclass[letterpaper]{article}
\usepackage{aaai}
\usepackage{mathptmx}
\usepackage{helvet}
\usepackage{amsmath}
\usepackage{amssymb}
\usepackage{graphicx}
\usepackage{subcaption}

\usepackage{xcolor}
\nocopyright

\begin{document}
\title{An Initial Attempt of Combining Visual Selective Attention \\ with Deep Reinforcement Learning}
\author{Liu Yuezhang\thanks{Work mostly performed at The University of Texas at Austin.\smallskip}\\
Department of Automation\\
Tsinghua University\\
\texttt{liuyuezhang15@mails.tsinghua.edu.cn}\\
\And
Ruohan Zhang\\
Department of Computer Science\\
The University of Texas at Austin\\
\texttt{zharu@utexas.edu}\\
\And
Dana H. Ballard\\
Department of Computer Science\\
The University of Texas at Austin\\
\texttt{danab@utexas.edu}\\
}

\maketitle
\begin{abstract}
\begin{quote}
Visual attention serves as a means of feature selection mechanism in the perceptual system. Motivated by Broadbent's leaky filter model of selective attention, we evaluate how such mechanism could be implemented and affect the learning process of deep reinforcement learning. We visualize and analyze the feature maps of DQN on a toy problem Catch, and propose an approach to combine visual selective attention with deep reinforcement learning. We experiment with optical flow-based attention and A2C on Atari games. Experiment results show that visual selective attention could lead to improvements in terms of sample efficiency on tested games. An intriguing relation between attention and batch normalization is also discovered.
\end{quote}
\end{abstract}

\section{Introduction}
In recent years, deep reinforcement learning (RL) has made significant progress in many domains, such as video games \cite{mnih_human-level_2015,hessel_rainbow:_2018}, the game of Go \cite{silver_mastering_2016,silver_mastering_2017} and continuous control tasks \cite{lillicrap_continuous_2015,gu_continuous_2016}. However, despite their impressive super-human end performance, many of the deep reinforcement learning methods require a large amount of samples for training. 

Visual attention, especially human gaze attention, has been widely studied in vision science. Broadbent proposed his famous filter model of selective attention in perceptual system~\cite{broadbent_perception_1958,treisman_monitoring_1964}. Research afterwards focused on modeling human gaze, based on the stimulus salience (bottom-up) \cite{itti1998model}, task priority (top-down) \cite{hayhoe_eye_2005}, or a combination of the both \cite{kanan_sun:_2009}. In these studies of gaze behaviors, it is conjectured that gaze serves as a feature selection mechanism based on saliency or task demand. Since nowadays feature selection is a core issue in machine learning, it is intuitive to study whether such mechanism could be beneficial for reinforcement learning since deep RL also requires feature extraction. Successful stories of combining attention with machine learning can be found in nature language processing \cite{vaswani_attention_2017}, computer vision \cite{palazzi_predicting_2018} and imitation learning \cite{zhang_agil:_2018}.

In this work, we propose a method to combine visual selective attention with deep reinforcement learning. We first analyze the feature maps of a deep neural network trained by reinforcement learning. Then we propose our method and evaluate on the toy problems as well as Atari games.

\section{Related Work}
\paragraph{Recurrent models for visual attention.}Several works implement visual attention using the glimpse network in combination with recurrent models \cite{larochelle_learning_2010,mnih_recurrent_2014}.  This approach was later extended to deep reinforcement learning in the Atari games domain, either using recurrent models \cite{sorokin_deep_2015} or glimpse sensory \cite{gregor_visual_2018}. However, these approaches only attempt to integrate visual attention at input level. Therefore, they resembled foveal vision at retinal level rather than visual selective attention, which might also exist in the deeper structure of the network. These pioneered works showed promising results which encourages us to further study how visual attention could benefit deep reinforcement learning. 
%In addition, due to the time reasons, all of the results reported in such work do not comply with today's reproducibility standard. 

\paragraph{Understanding deep RL.} Understanding features learned by deep RL is critical for our goal since we target at combining attention with learned features. Some research visualize and analyze the property of the learned policy by deep Q-network (DQN)~\cite{mnih_human-level_2015} based on t-SNE and SAMDP \cite{zahavy_graying_2016}. Others use perturbation to extract the visual features the agent attended to~\cite{greydanus_visualizing_2018}. However, most of these work focused on either the input layer or the last several fully connected layers. Thus, the feature maps in the mid layers and their relation with visual inputs and task reward requires further investigation.

\paragraph{Combining different sources of knowledge in CNN.} Hypothetically, introducing visual attention into deep RL can be thought as integrating a different source of visual information into an existing neural network. Similar problems have been studied extensively by the computer vision community. For example, multiplicative fusion with optical flow in CNN is proposed for action recognition \cite{park_combining_2016}. More fusion methods in two-stream network were discussed \cite{feichtenhofer_convolutional_2016}. As for video prediction, action-conditional architecture also involve a multiplication structure to combine action information \cite{oh_action-conditional_2015}. %However, all of the work did not extend to deep reinforcement learning domain, and the critical relationship between combination and normalization was not discovered.

\section{The Toy Problem Catch}
\begin{figure}
    \centering
    \includegraphics[width=\linewidth]{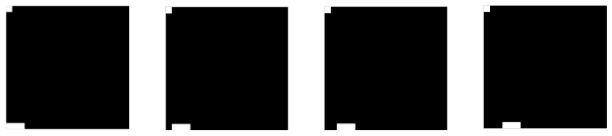}
    \caption{The Catch environment. A set of visual input for evaluating the learning process is shown above (4 out of 18). In this set, the ball is fixed at the up left corner, while the paddle moves from the bottom left to right.}
    \label{fig:eval}
\end{figure}

A toy problem called Catch was designed to study the properties of the CNN feature maps in deep reinforcement learning (see Figure \ref{fig:eval}). A ball falls down from the top of the screen, and the agent controls the paddle at the bottom to catch the ball. To be more specific:
\begin{itemize}
    \item State space: $20\times20$, with only black and white $(0/1)$ pixels. The ball is of size $1\times1$ and the paddle is $3\times1$.
    \item Initial state: at the beginning of each episode, the ball and the paddle are placed at a random position at the top and the bottom row respectively.
    \item Action space: $3$ actions, move the paddle 1 pixel left, stay, or move right.
    \item Dynamics: the ball falls at the speed of 1 pixel per timestep.
    \item Episode length: each episode ends when the ball reaches the bottom, thus it is fixed to 20 timesteps.
    \item Reward: the player only receives reward at the end of episodes, $+1$ if the paddle catches the ball successfully, $0$ otherwise. 
\end{itemize}

\section{Understanding Feature Maps in DQN}
Visual attention can be treated as a special visual input that highlights important spatial information. Such additional visual input, or mask, could be applied at input level~\cite{zhang_agil:_2018} or at higher level, i.e., deeper in the network~\cite{park_combining_2016,feichtenhofer_convolutional_2016}. We are interested in the effect of introducing attention at the last convolution layer of DQN. Therefore we first need to examine the features learned by the last convolution layer of DQN to ensure that it is sensible to combine attention with these features. 

To achieve this goal, we choose the classic DQN as our learning algorithm for Catch~\cite{mnih_human-level_2015}. We preserve the technical details of DQN as much as possible except for three modifications. Since the observation space reduces from $84\times84\times4$ to $20\times20\times1$, we simply remove the first convolution layer of DQN, as the size of the second convolution layer in original DQN-CNN is exactly $20\times20$. We use Adam optimizer here, rather than RMSProp, with a learning rate $\alpha=5e-4$. As for the hyperparameters, we modify several hyperparameters as the game is simple.

The convolution layers of DQN generally serve as a visual feature extractor. In the Catch environment, according to the modifications we mention above, the size of the input and feature map at each layer are $(20, 20, 1)$, $(9, 9, 64)$ and $(7, 7, 64)$ respectively. We now visualize the feature maps of the last convolution layer. 

\begin{figure}
    \centering
    \includegraphics[width=\linewidth]{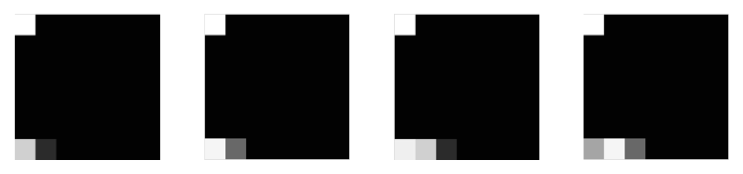}
    \caption{Averaged feature maps of the last convolution layer (4 out of 18), corresponding to Figure \ref{fig:eval}.}
    \label{fig:avr}
\end{figure}

A fixed set of visual stimulus is made to evaluate the learning process of DQN, as shown in Figure \ref{fig:eval}. The stimulus set is feed to the trained DQN. We then record the outputs of feature maps from the last convolution layer. One way to visualize these feature maps is to directly average all the feature maps, as shown in Figure \ref{fig:avr}. Notice that these averaged feature maps are very similar to the visual inputs shown in Figure~\ref{fig:eval}. They all have a 'ball' at the upper left corner, and a 'paddle' at the bottom. 

%In our case, the feature maps of convolution layers preserve the spatial information of visual inputs. 

\begin{figure}
    \centering
    \begin{subfigure}{.25\textwidth}
        \centering
        \includegraphics[width=.8\linewidth]{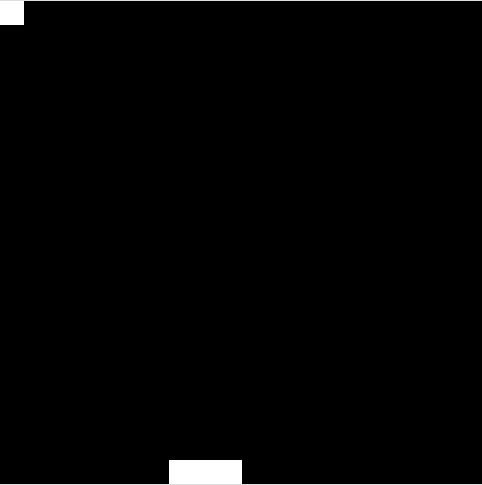}
        \caption{The visual input}
        \label{fig:input_6}
    \end{subfigure}%
    \begin{subfigure}{.25\textwidth}
        \centering
        \includegraphics[width=.8\linewidth]{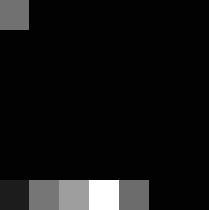}
        \caption{No manipulation}
        \label{fig:conv_avr_6}
    \end{subfigure}
    \begin{subfigure}{.25\textwidth}
        \centering
        \includegraphics[width=.8\linewidth]{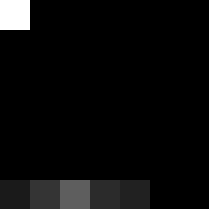}
        \caption{$\gamma=0$}
        \label{fig:gamma=0}
    \end{subfigure}%
    \begin{subfigure}{.25\textwidth}
        \centering
        \includegraphics[width=.8\linewidth]{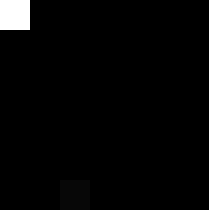}
        \caption{$r=0$}
        \label{fig:r=0}
    \end{subfigure}
    \caption{(a) Visual input provided to the trained DQN. (b) Averaged feature map of the last convolution layer. (c) Averaged feature map of the last convolution layer when setting $\gamma=0$. (d) Averaged feature map of the last convolution layer when setting $r=0$.}
    \label{fig:Catch_revised}
\end{figure}

We hypothesize that the blurred paddle-like object in averaged feature maps may represent a distribution related to the position of the paddle, rather than the exact shape of the paddle, where the former is closely related to the task reward. To test this hypothesis, we manipulate reward-related variables (reward and discount factor) to further study the potential relation between feature maps and task reward. The results are shown in Figure \ref{fig:Catch_revised}, where (b) visualizes the averaged feature map learned with the original Catch environment. 

\medskip
\textbf{Zero Discounted factor.} If the discounted factor is set to 0, the intensity of the paddle object significantly decreases as shown in \ref{fig:Catch_revised} (c). 

\medskip
\textbf{No reward.} Furthermore, we set the reward to 0, which means the agent would not receive any reward from this environment. Then the paddle object completely disappeared as shown in (d). 

\medskip
These findings are consistent with previous results that the features learned by a neural network is modulated by task-related variables~\cite{gluck1993hippocampal}. We conclude that the feature maps learned by DQN is a joint representation of both visual inputs and task reward. Visual spatial information is preserved through convolution layers, and the intensity of the feature maps is associated with the reward. Therefore, applying visual attention at the last convolution layer is a reasonable choice. 

\section{Combining Visual Attention with Deep RL}
\begin{figure*}
    \centering
    \includegraphics[width=\linewidth]{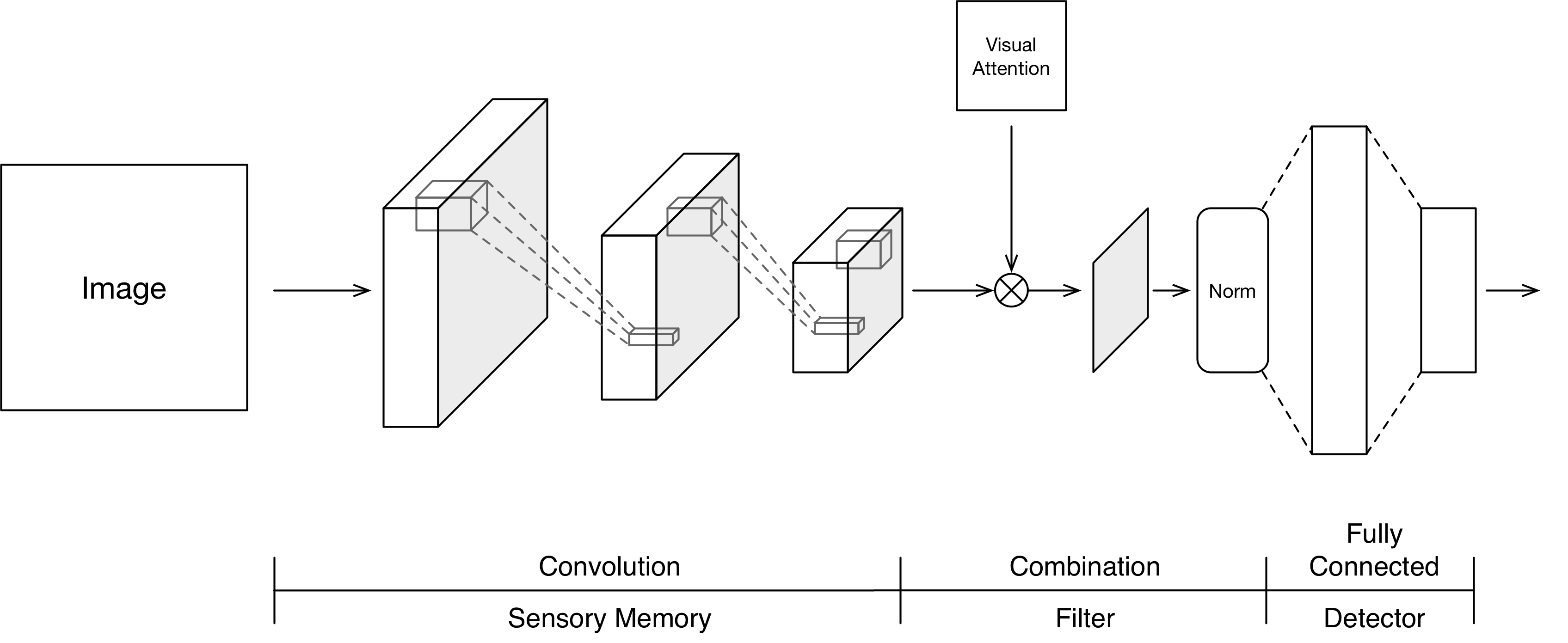}
    \caption{The architecture of combining visual selective attention with deep reinforcement learning. The pipeline illustrates the original Broadbent's leaky filter model of selective attention (see Figure \ref{fig:broadbent}).}
    \label{fig:adqn}
\end{figure*}

We now propose a method to combine visual selective attention with deep reinforcement learning, as shown in Figure~\ref{fig:adqn}.

 The feature maps of size $k \times l$ with $c$ channels $\boldsymbol{h}^{sense}_t \in \mathbb{R}^{k \times l \times c}$ at time $t$  are:
\begin{equation}
    \boldsymbol{h}^{sense}_{t}=CNN(\boldsymbol{x}_{t-m+1:t})\label{eq:sense}
\end{equation}
where $\boldsymbol{x}_{t-m+1:t} \in \mathbb{R}^{h \times w \times m}$ denotes last $m$ frames of visual inputs with size $h \times w$. 

Let $\phi$ denotes some function generating a visual attention map $\boldsymbol{g}_{t} \in \mathbb{R}^{h \times w}$ based on the visual inputs at time $t$:
\begin{equation}
    \boldsymbol{g}_{t}=\phi(\boldsymbol{x}_{t-m+1:t})\label{eq:att}
\end{equation}
We first reshape the attention mask $\boldsymbol{g}_{t}$ to be the same shape as the feature maps $\boldsymbol{h}^{sense}_{t}$. The values in the attention mask are in the range $[1,\theta]$ instead of an ordinary $[0,1]$ mask. Hence no information is discarded but more important features are scaled up. In this paper, we set $\theta=2$.

Next, we simply multiply the attention mask with the feature maps of the last convolution layer elementwise:
\begin{equation}
    \boldsymbol{h}^{att}_{t}=\boldsymbol{h}^{sense}_{t}\odot Rescale(Reshape(\boldsymbol{a}_t), \theta)\label{eq:mul}
\end{equation}

Moreover, the scale of the feature maps is also critical for deep reinforcement learning. Therefore we add a batch normalization \cite{ioffe_batch_2015} layer after the multiplication operation:
\begin{equation}
    \boldsymbol{h}^{filt}_{t}=Norm(\boldsymbol{h}^{att}_{t})\label{eq:norm}
\end{equation}
The filtered feature vector $\boldsymbol{h}^{filt}_{t}$ is then fed to the fully connected layers in DQN. Such normalization structure could be understood as a naive brightness adaptation mechanism \cite{purves_vision:_2004}, and is also proposed in other attention models \cite{vaswani_attention_2017}.

\section{Experiments and Result}
Given the techniques proposed, we experiment using DQN with or without attention and batch normalization. We first combine hand-crafted attention with DQN \cite{mnih_human-level_2015} on the Catch environments as a sanity check. After that we extend to Atari Games in which we combine optical flow-based attention with A2C \cite{mnih_asynchronous_2016}. Due to the stochastic nature of the training process and in compliance with the reproducibility standard~\cite{henderson_deep_2018}, all experiments are repeated with different random seeds.

\subsection{Experiments with Catch}
\begin{figure}
    \centering
    \begin{subfigure}{.25\textwidth}
        \centering
        \includegraphics[width=.8\linewidth]{input_7.png}
        \caption{The original Catch}
        \label{fig:catch_vs_origin}
    \end{subfigure}%
    \begin{subfigure}{.25\textwidth}
        \centering
        \includegraphics[width=.8\linewidth]{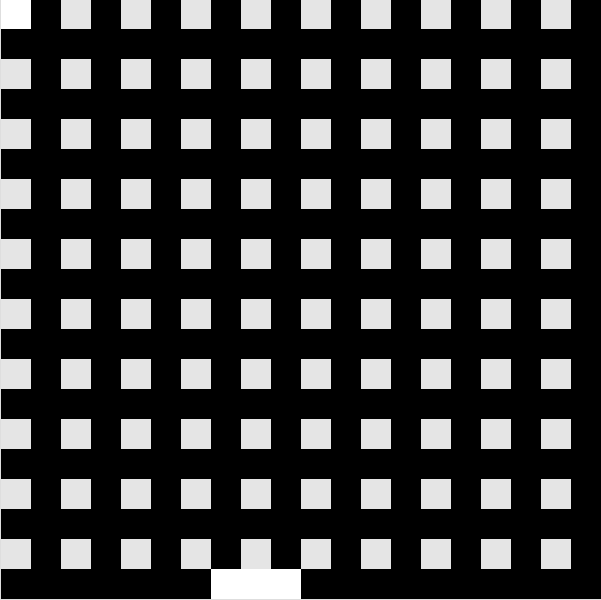}
        \caption{Catch with background}
        \label{fig:catch_vs_bg}
    \end{subfigure}
    \caption{Catch vs Catch with background.}
    \label{fig:catch_vs}
\end{figure}

\begin{figure*}
    \centering
    \begin{subfigure}{.5\textwidth}
        \centering
        \includegraphics[width=\linewidth]{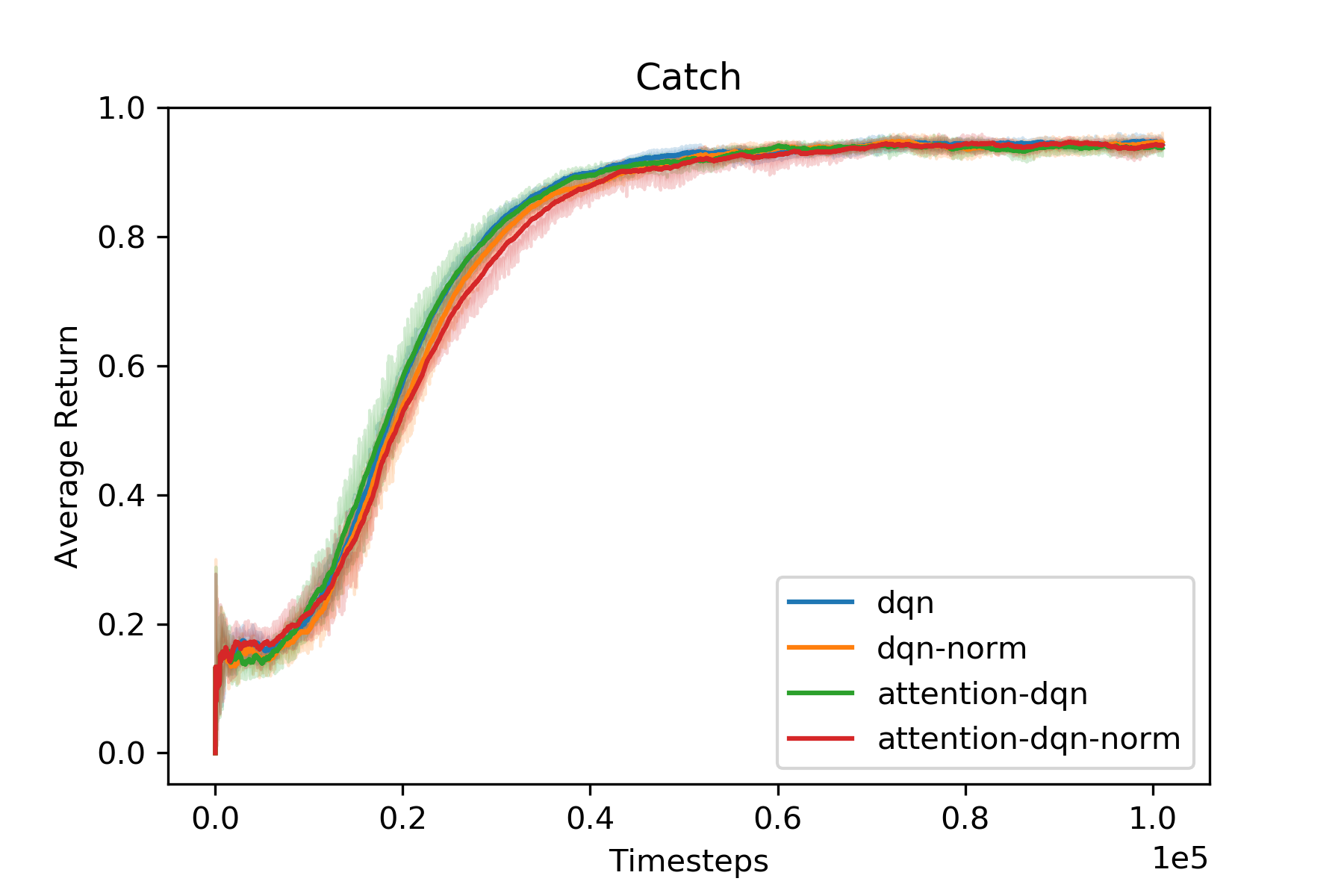}
        \caption{Catch}
        \label{fig:res_catch_catch}
    \end{subfigure}%
    \begin{subfigure}{.5\textwidth}
        \centering
        \includegraphics[width=\linewidth]{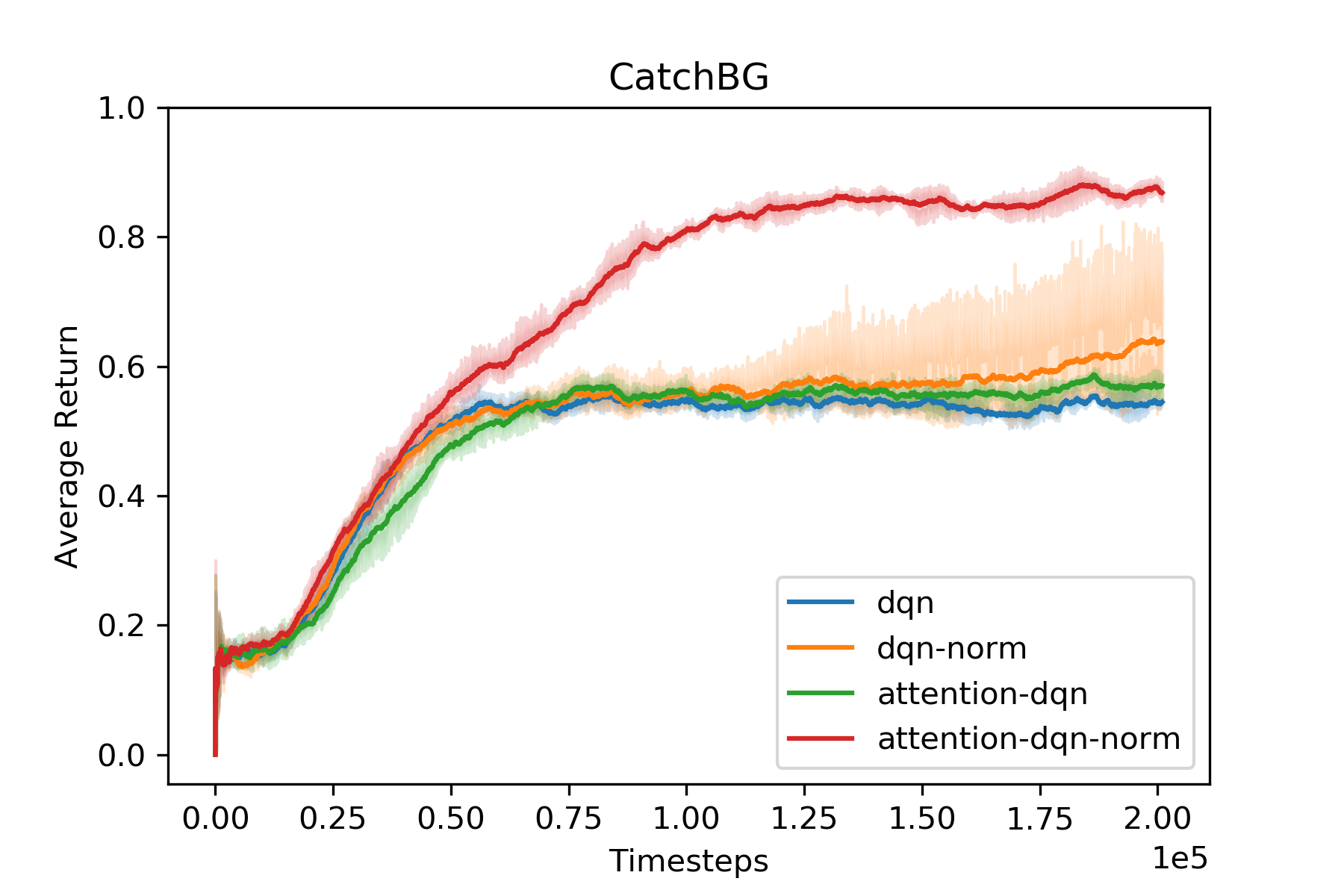}
        \caption{Catch with Background}
        \label{fig:res_catch_catchbg}
    \end{subfigure}
    \caption{Average reward over last 100 episodes for Catch with different learning methods. 'opt-a2c-norm' denotes A2C with optical flow-based attention and batch normalization. Experiments are repeated with 5 random seeds and the shaded region denotes 95\% confidence interval.}
    \label{fig:res_catch}
\end{figure*}

\begin{figure*}
    \centering
    \begin{subfigure}{.5\textwidth}
        \centering
        \includegraphics[width=\linewidth]{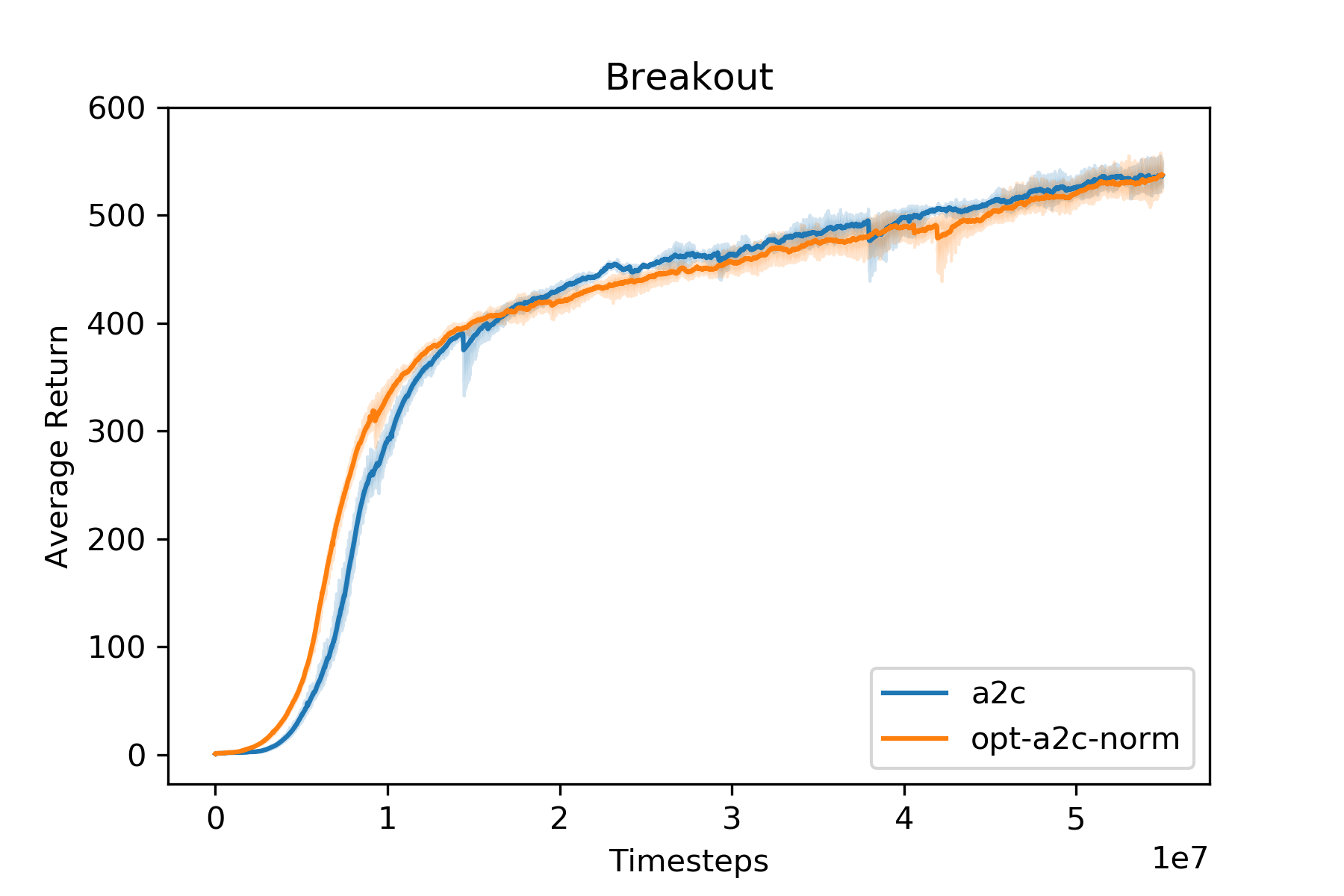}
        \caption{Breakout}
        \label{fig:res_atari_breakout}
    \end{subfigure}%
    \begin{subfigure}{.5\textwidth}
        \centering
        \includegraphics[width=\linewidth]{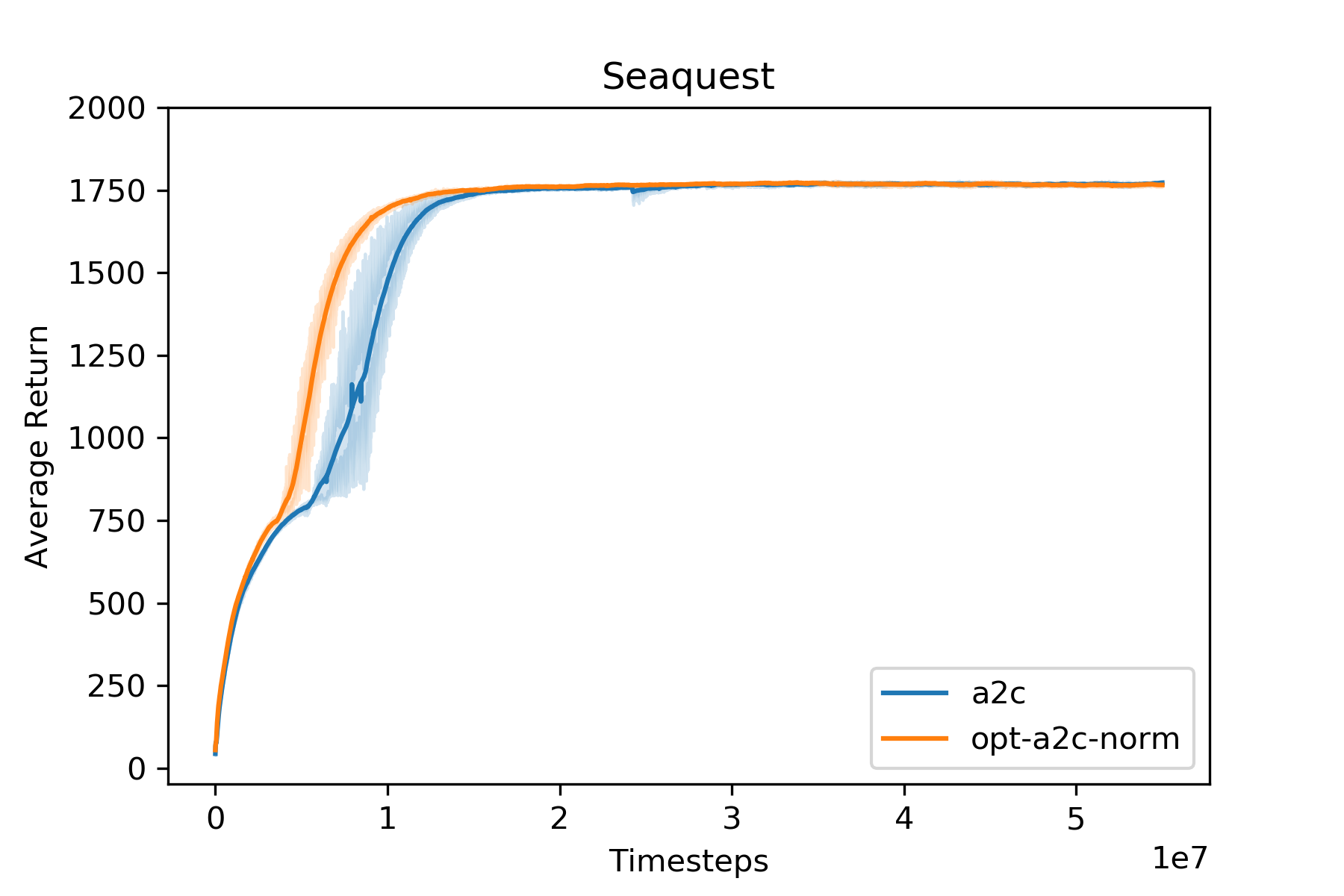}
        \caption{Seaquest}
        \label{fig:res_atari_seaquest}
    \end{subfigure}
    \begin{subfigure}{.5\textwidth}
        \centering
        \includegraphics[width=\linewidth]{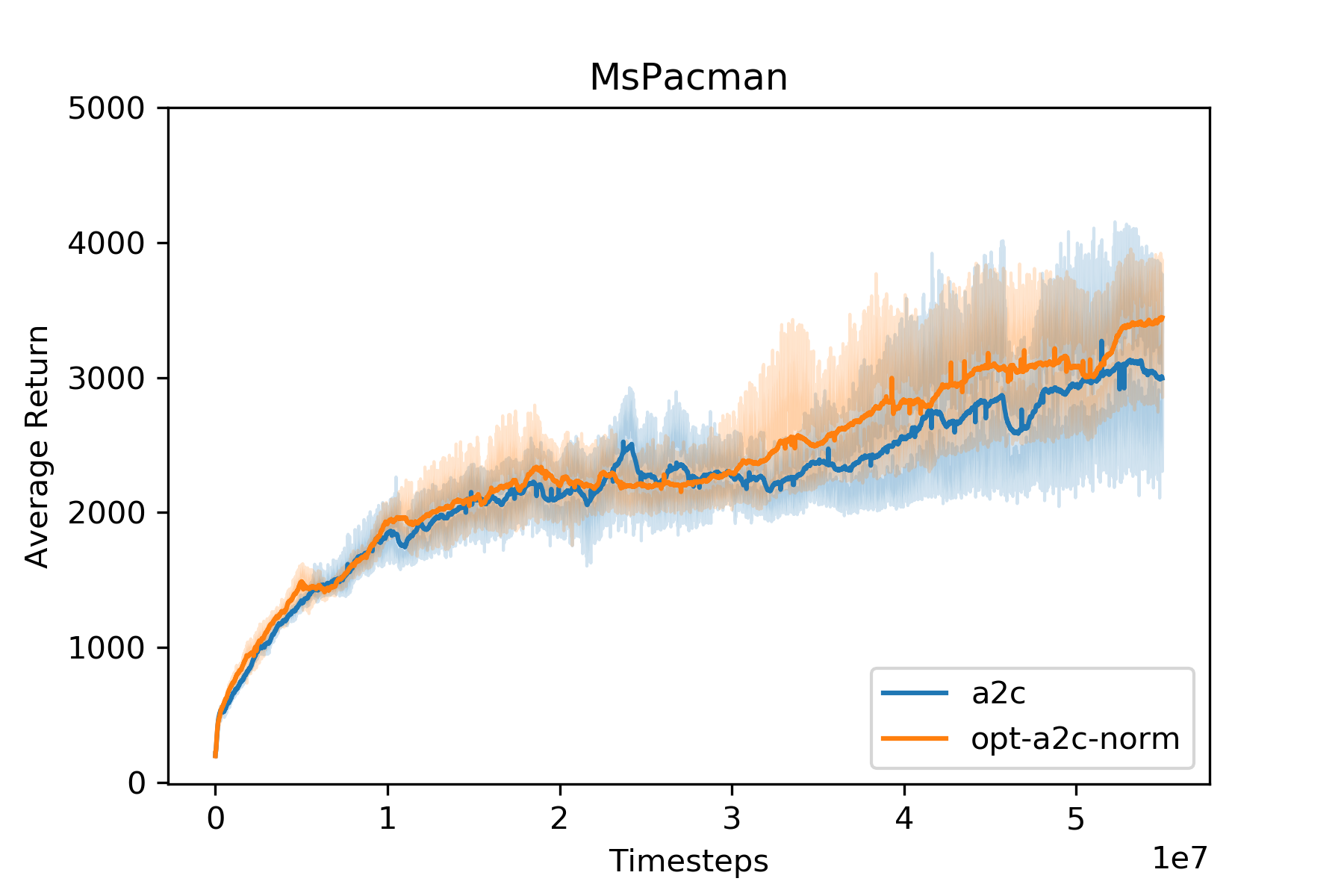}
        \caption{MsPacman}
        \label{fig:res_atari_mspacman}
    \end{subfigure}%
    \begin{subfigure}{.5\textwidth}
        \centering
        \includegraphics[width=\linewidth]{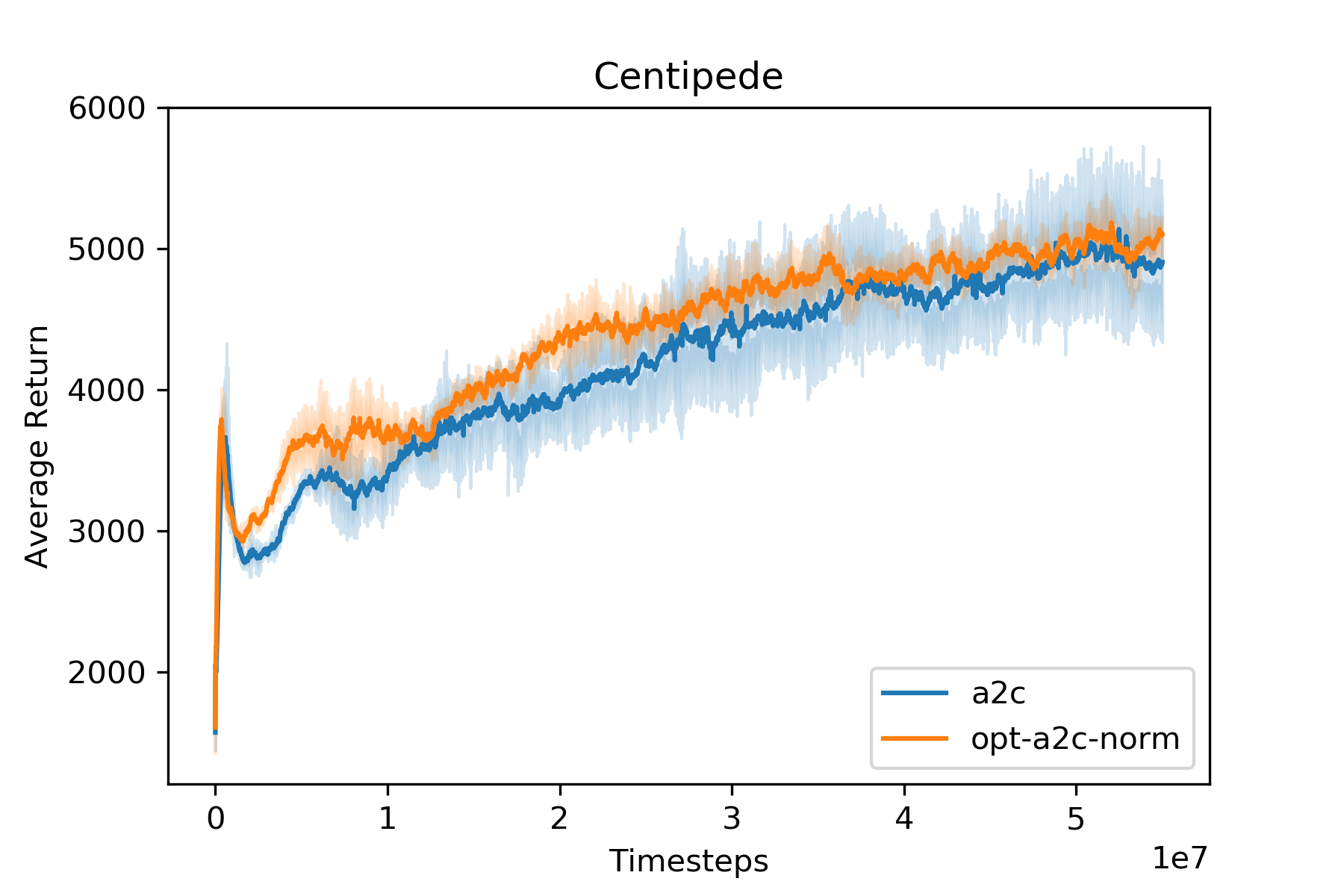}
        \caption{Centipede}
        \label{fig:res_atari_centipede}
    \end{subfigure}
    \caption{Average reward over last 100 episodes for four Atari games with different learning methods. 'opt-a2c-norm' denotes A2C with optical flow-based attention and batch normalization. Experiments are repeated with 5 random seeds and the shaded region denotes 95\% confidence interval. }
    \label{fig:res_atari}
\end{figure*}

The effect of attention may not be very salient given a simple visual task with clean inputs. Inspired by~\cite{mnih_recurrent_2014}, we modify the original Catch environment to include certain level of background noise. A latticed background which has a pixel value of 0.9 at every odd coordinate position is added upon the original image. We name this environment 'Catch with Background', or simply 'CatchBG' (see Figure \ref{fig:catch_vs}).

\begin{figure*}
    \centering
    \includegraphics[width=\linewidth]{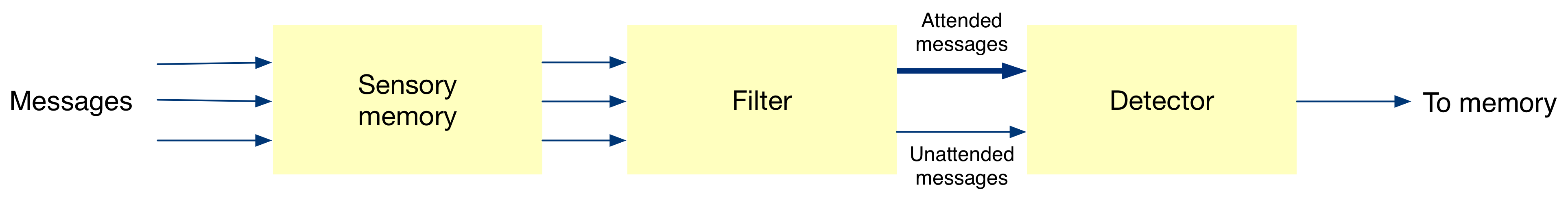}
    \caption{Broadbent's leaky filter model of selective attention. Modified from the illustration \cite{goldstein_attention_2014}.}
    \label{fig:broadbent}
\end{figure*}

CatchBG could be a lot more difficult for deep reinforcement learning methods without visual attention, since the agent would observe a screen full of 'balls' to catch - the agent is forced to learn to tell the difference between 0.9 and 1 in grayscale. %, or the subtraction operation, which could be a quite challenging task for neural networks \cite{He_2016_CVPR}.

Here, we use hand-crafted features as visual attention. The attention map marks the true position of the ball and the paddle. So it is identical to the visual inputs of the original Catch environments. We downsample the attention map by maxpooling to match the size of the feature maps, rescale it and combine with DQN by multiplication. Results comparing different methods on the Catch environments are shown in Figure \ref{fig:res_catch}.

In the Catch environment, all four methods perform similarly. All methods get a more than 0.9 averaged return (which also denotes an over 90\% successful rate since we average along past 100 episodes and set the final exploration rate $\epsilon=0.1$) after 40000 timesteps, indicating that DQN has learned the task successfully.

However, for the CatchBG environment, only the method with both visual attention and batch normalization can still achieve a similar performance - over 90\% success rate. The other three methods can only success in about 50\% of the trials -- which is exactly the probability that the ball is initialized in the columns with no background, as we only place background at the odd coordinate grids - therefore, fail to learn the difference between the ball and the background in this environment. In addition, attention alone without normalization also results in a poor performance in this test. 
%This interesting phenomenon confirms our two conclusions proposed by visualizing and understanding the feature maps in the section above.

\subsection{Experiments with Atari Games}
For more complex visuomotor tasks such as Atari games, a reasonable source of attention is motion, as objects movement are good indicator of important visual features in video games. Hence we use optical flow between two frames~\cite{farneback2003two} to construct an attention map. We combine optical flow-based attention with A2C on Atari games following the same method and test our approach on four popular Atari games - Breakout, Seaquest, MsPacman and Centipede. Results are shown in Figure \ref{fig:res_atari}. We use OpenAI baselines \cite{baselines} with default hyperparameters.

Comparing with the original A2C algorithm, moderate performance improvements are observed at different stages in all four tested games. Improvements at early stage of learning are present in Breakout and Seaquest, since moving objects are critical for both of the games, which indicates that the optical flow might be an appropriate attention source. Also the rules of the games are quite straightforward, therefore improvements in sensory system are more likely to result in improvements in task performance. MsPacman and Centipede are games with many visually identical objects, hence motion-based attention is likely to help to tell apart the objects that matters for the current decision step. Experiments on more games should be conduct to provide a more comprehensive evaluation for the effect of introducing visual attention.
%However, the bottleneck of deep reinforcement learning might mainly presence in the reinforcement learning part, therefore improvements from sensory system are not guaranteed to result in significant improvements in task performance. 

\section{Discussion}
Before drawing the conclusion, we would like to discuss the potential relation between our approach and Broadbent's leaky filter model of selective attention \cite{broadbent_perception_1958}. The model is shown in Figure \ref{fig:broadbent}.

In Broadbent's filter model of selective attention, messages flow through the sensory memory, filter and detector successively~\cite{broadbent_perception_1958}. In our model, these three components roughly correspond to the convolution layers, integration (multiplication and normalization) and fully connected layers in the network. The sensory memory holds the input messages for a short period, extracts features and passes to the filter. The filter then selects the attended messages based on their physical characteristics, as we choose optical flow for instance, and filters out the others. The attended messages are then passed to the detector for further cognitive processes \cite{broadbent_perception_1958}.

We should note that in the original Broadbent's filter model, only attended messages could pass through the filter. However, a modification was proposed based on several psychology experiments, which suggested that unattended messages are also able to pass through the filter \cite{treisman_monitoring_1964}. It is known as leaky filter model. Therefore, we multiply our visual attention scaling from 1 rather than 0 to simply keep all the unattended messages.

A significant difference still exists between the sensory memory module and convolution layers, since CNNs do not explicitly model memory. It could be an interesting topic to study how sensory memory contribute to the generation of visual attention function $\phi$. In addition, unlike the visual scanning that could only exist at the input layer, we presume that the selective attention mechanism might be universal as in the neural gain model \cite{eldar_effects_2013}, and could be utilized hierarchically in any layer. 

%Predictive coding could also be another intriguing perspective \cite{rao_predictive_1999}. Note: if mention predictive coding better to cite relevant work in RL

\section{Conclusion and Future Work}
In this paper, we aim at combining visual selective attention with deep reinforcement learning. We first try to understand the feature maps on the toy problem Catch, and then propose our approach for the fusion problem. An interesting interaction effect between attention and normalization is discovered and should be studied more carefully. The new learning method achieves performance improvements on tested games. The simple CatchBG example demonstrates that visual attention could be particularly helpful in an environment with cluttered and noisy visual input. 

However, many issues remain to be solved to achieve a potentially better performance. For example, the attention map is approximated by optical flow, which can be replaced with a more comprehensive saliency model~\cite{itti1998model}, or a real human attention model from gaze~\cite{zhang_agil:_2018}, or generated by the sensory memory as mentioned in the discussion. It's also possible that the first several convolution layers have stronger correlation with input, while the last several layers correlates more with the value function. Hence at which layer (or probably all layers respectively) attention should be integrated to produce the best performance could also be explored. 

\section{Acknowledgments}
The work was supported by NIH Grant EY05729, NIH
Grant T32 EY21462-6, NSF Grant CNS-1624378, Google AR/VR Research Award and Tsinghua Initiative Scientific Research Program.

\bibliographystyle{aaai}
\bibliography{adrl}

\end{document}